\documentclass[10pt]{cai}
\usepackage[table]{xcolor}
\usepackage{float}
\usepackage{graphicx}
\usepackage{subcaption}

\begin{document}
\def\conferenceyear{2026}
\volumeheader{37}{0}
\begin{center}

\title{Differentiable Logic Gate Networks for Low-Latency EEG Classification on Edge Devices}
\maketitle

\thispagestyle{empty}

\begin{tabular}{cc}
Shyamal Y. Dharia\upstairs{\affilone}, Stephen D. Smith\upstairs{\affilone,\affiltwo}, Camilo E. Valderrama\upstairs{\affilone, \affilthree, \affilfour *}
\\[0.25ex]
{\small \upstairs{\affilone} The University of Winnipeg, Department of Applied Computer Science} \\
{\small \upstairs{\affiltwo} The University of Winnipeg, Department of Psychology} \\
{\small \upstairs{\affilthree} University of Manitoba, Department of Electrical and Computer Engineering} \\
{\small \upstairs{\affilfour} University of Calgary, Department of Community Health Sciences } \\
\end{tabular}
  
\emails{
  \upstairs{*}c.valderrama.uwinnipeg.ca 
  
}


\end{center}

\begin{abstract} 

Real-time EEG classification on edge devices is bottlenecked by the floating-point arithmetic of conventional neural networks. We investigated Differentiable Logic Gate Networks (Diff-Logic) as a hardware-native alternative that compiles models into pure Boolean circuits executable via bitwise CPU operations. Through rigorous iso-parameter experiments across four EEG datasets spanning two classification tasks, binary dementia detection and 3-class emotion recognition, we compared Diff-Logic against matched-capacity Multi-Layer Perceptron (MLP) and Binarized Neural Network (BNN) baselines at four complexity tiers (50k--500k parameters). On dementia screening, Diff-Logic achieved 80.2\% Macro F1, outperforming the MLP baseline by 6.8\%. On emotion recognition, the MLP retained a moderate performance advantage but incurred a 2.3$\times$ higher latency and 14$\times$ larger model size when deployed on a power-constrained (7W) Nvidia Jetson Orin Nano CPU (Single-core). Critically, Diff-Logic inference time remained nearly constant across a 10$\times$ increase in model scale, achieving a peak speedup of 2.9$\times$ over MLPs at the largest complexity tier. Our results establish logic-based neural architectures as a practical paradigm for resource-constrained brain--computer interfaces, achieving competitive or superior performance while natively satisfying the latency and memory constraints of portable edge deployment. Code is available on GitHub: \url{https://github.com/Shyamal-Dharia/eeg-difflogic}.

\end{abstract}

\copyrightnotice

\section{Introduction}
\label{sec:introduction}

Although Artificial Neural Networks (ANNs) have revolutionized pattern recognition in complex biosignals~\cite{Raza2025, lai2025simple}, their evolution toward increasingly high-parameter architectures has created a significant deployment barrier~\cite{Joel2024, kuruppu2025eeg}. This barrier is present, for instance, in the brain-computer interface (BCIs) and real-time EEG monitoring, where the computational overhead of standard deep learning models often exceeds the power and thermal envelopes of portable edge devices. Consequently, practical deployments are often forced to offload processing to the cloud, thereby introducing latency that disrupts the real-time feedback loop essential for BCIs while exposing sensitive neural data to security risks \cite{Muter2025}.



To address the latency limitations of deployed deep learning models in real-time BCI systems, it is essential to move away from traditional multiply-accumulate (MAC) operations toward hardware-native formulations capable of delivering high-speed inference within microsecond-scale budgets~\cite{sen2025low}. One promising approach in this direction is Differentiable Logic Gate Networks (Diff-Logic) ~\cite{petersen2022deep, petersen2024convolutional, buhrer2025recurrent, ruttgers2025light}. By relaxing discrete Boolean operations, such as approximating AND gates through the product of binary inputs, researchers have enabled the training of logic-based models via traditional gradient descent. Unlike conventional networks that rely on high-precision floating-point weights, these models converge into a hard, sparse network of Boolean gates. This transition from floating-point arithmetic to bitwise logic allows for extreme efficiency, as the resulting model can be executed directly on the Arithmetic Logic Unit (ALU) of edge processors, bypassing the energy-intensive floating-point bottlenecks of conventional AI accelerators.


Although Diff-Logic shows competitive accuracy on standard benchmarks, its application to real-world EEG processing under edge constraints remains unexplored. We investigate whether its theoretical efficiency translates to complex, non-stationary brain signals. 

In this paper, we present the first rigorous application of Diff-Logic to the constraints of real-time EEG signal decoding. To this end, we investigate how bitwise learning dynamics respond to the high-dimensional, non-stationary nature of brain signals, comparing them with traditionally floating-point intensive models used for an EEG-based classification task. Our main contributions are summarized below:


\begin{itemize} 

\item We show that Diff-Logic effectively encodes high-dimensional EEG features, matching or outperforming floating-point baselines in emotion recognition and dementia detection.

\item We propose a pipeline from soft probabilistic training to compiled Boolean inference, achieving up to a $2.91\times$ speedup over Float32 MLPs on embedded hardware. 

\item We provide an iso-parameter evaluation across four EEG benchmarks (50k–500k parameters), comparing Diff-Logic against MLPs and BNNs to characterize performance-efficiency trade-offs. 
\end{itemize}





\section{Related Works}

\subsection{Deep Learning in EEG-based Brain-Computer Interfaces}

The evolution of EEG decoding has transitioned from traditional feature engineering to deep representation learning. Early successes were marked by architectures such as Deep4Net~\cite{schirrmeister2017deep}, which used deep convolutional networks to capture spatiotemporal hierarchies in brain signals. However, the high parameter count of such models posed challenges for real-time deployment. This led to the development of compact architectures, most notably EEGNet~\cite{lawhern2018eegnet}, which introduced depthwise and separable convolutions to significantly reduce the computational burden while maintaining competitive accuracy across diverse BCI tasks. Despite these optimizations, recent trends have shifted toward even more complex structures, such as EEG-Conformers~\cite{song2022eeg}, which leverage self-attention mechanisms to model long-range temporal dependencies. Furthermore, the emergence of EEG foundation models, such as LaBraM~\cite{jiang2024large}, BIOT~\cite{yang2023biot}, and the recent DeeperBrain~\cite{wang2026deeperbrain}, has introduced a new paradigm of large-scale pre-training. While these models offer unprecedented generalization across subjects and tasks, they are highly parametric, often requiring millions to billions of floating-point operations (FLOPs) per inference. Even lightweight foundation models using Mamba-based linear scaling, such as LUNA~\cite{doner2025luna} and FEMBA~\cite{tegon2025femba}, remain fundamentally dependent on high-precision floating-point arithmetic. Although knowledge distillation~\cite{hinton2015distilling} and structured pruning~\cite{han2015deep} offer partial remedies, they retain a fundamental dependence on MAC operations, thereby creating a persistent thermal and energy bottleneck for battery-powered wearable devices and leaving a critical efficiency gap for portable edge deployment.

\subsection{Differentiable Logic and Hardware-Native AI}
The pursuit of hardware-efficient AI has historically focused on quantization and pruning. Binarized Neural Networks (BNNs)~\cite{courbariaux2016binarized} represented a significant leap, constraining weights and activations to $\{-1, +1\}$ and replacing multiplications with XNOR-popcount operations. However, BNNs retain the dense connectivity of standard deep learning architectures, requiring accumulation steps that limit throughput on embedded BCI platforms. A distinct paradigm has emerged through the work of Petersen et al.~\cite{petersen2022deep}, who introduced \textit{Differentiable Logic Gate Networks}. Unlike BNNs, which quantize a fixed topology, this framework learns the topology itself. By relaxing Boolean logic gates (AND, OR, XOR) into continuous differentiable operators, these networks can be trained via gradient descent to identify sparse, optimal logic circuits. Upon convergence, each gate hardens to a single Boolean function, yielding a network that requires no accumulation registers and executes entirely through bitwise logic. Recent advancements have extended this framework to convolutional structures~\cite{petersen2024convolutional} and recurrent units~\cite{buhrer2025recurrent}, demonstrating efficacy on tabular and image datasets (e.g., MNIST, CIFAR). However, the application of diff-logic to continuous, non-stationary physiological time-series remains unexplored. While BNNs have been applied to EEG~\cite{mirsalari2020mubinn}, true logic gate networks, which eliminate accumulation in favour of pure bitwise operations, have not yet been rigorously evaluated for neurophysiological signal classification. This gap is particularly critical for edge computing platforms, such as the NVIDIA Jetson, where efficient CPU-bound execution is required to decouple the BCI control loop from AI accelerators, thereby reserving the GPU for concurrent, computationally intensive downstream workloads.



\section{Methods}

\begin{figure*}[t]
    \centering
    \includegraphics[width=1\linewidth]{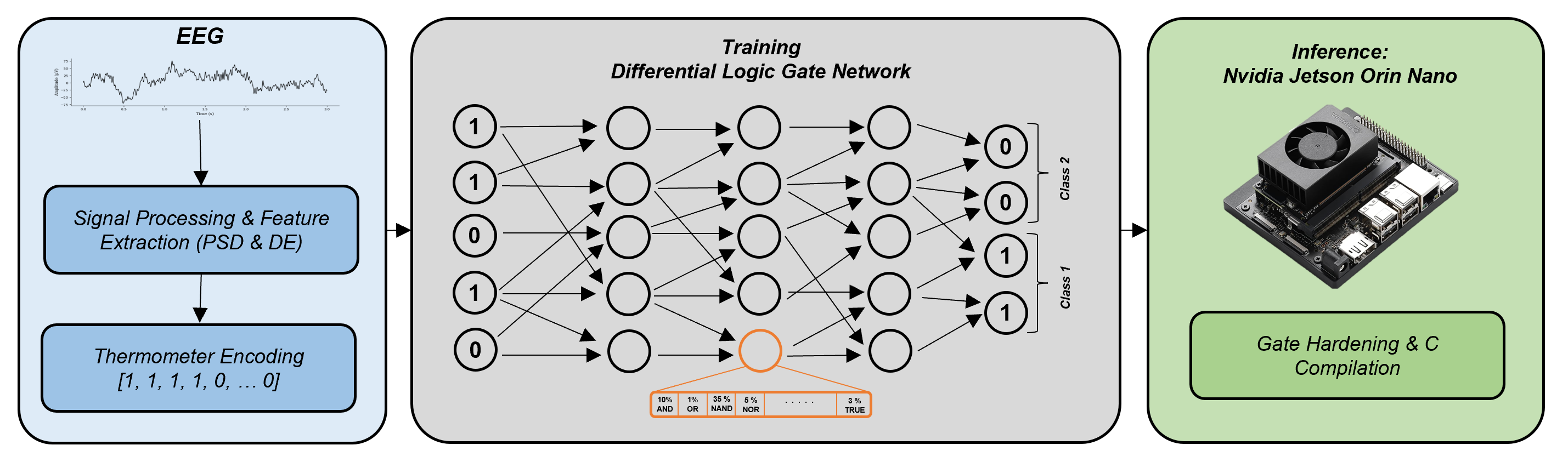}
    \caption{Proposed workflow for EEG classification using Diff-Logic.}
    \label{fig:workflow}
    \vspace{-0.5cm}
\end{figure*}


\subsection{Dataset and Feature Extraction}
To evaluate the robustness of Diff-Logic across clinical and affective domains, we utilized four EEG datasets spanning two distinct EEG-based applications. In total, the proposed framework was evaluated across a diverse cohort of 119 unique subjects (88 from the clinical dataset and 31 from SEEDs; see Table~\ref{tab:datasets}).

\subsubsection{Clinical Diagnosis: Dementia}
The dementia dataset~\cite{miltiadous2023dataset} focuses on detecting neurodegenerative diseases using resting-state EEG from 88 subjects: 29 Cognitively Normal (CN), 36 Alzheimer's Disease (AD), and 23 Frontotemporal Dementia (FTD). Signals were recorded at 500\,Hz using a 19-channel montage (10--20 system). We extracted Power Spectral Density (PSD) features across five frequency bands ($\delta, \theta, \alpha, \beta, \gamma$), yielding a feature vector of $19 \times 5 = 95$ dimensions per 10-second non-overlapping window. The analysis was divided into two binary classification tasks: CN vs.\ AD (65 subjects) and CN vs.\ FTD (52 subjects). To ensure clinical validity, we employed subject-level 10-fold cross-validation, ensuring that no windows from the same subject appear in both training and testing partitions. Final per-subject predictions were obtained via majority voting across all constituent windows.

\subsubsection{Emotion Recognition: SEED Family}
We utilized the SEED dataset~\cite{zheng2015investigating} and its cross-cultural variants (France and Germany)~\cite{liu2022identifying} for three-class emotion recognition (Negative, Neutral, Positive). All variants used a 62-channel setup downsampled at 200\,Hz, segmented into 4-second non-overlapping windows. The primary feature was Differential Entropy (DE) across five bands, yielding a $62 \times 5 = 310$ dimensional vector per window. These datasets were evaluated using trial-level 10-fold cross-validation, where each trial corresponds to a specific film clip viewed during a session. This approach is subject-dependent, where the model learns from a subset of a subject's trials to predict emotional state in held-out trials. Unlike clinical diagnosis, where generalization to unseen patients is critical, emotion recognition systems for BCIs often benefit from an initial on-device calibration phase~\cite{lotte2018review}. Our protocol, therefore, mimics a practical deployment scenario where the model is personalized to a specific user's EEG.


\subsection{Per-Fold Feature Normalization}
To ensure robust scaling of PSD and DE features without data leakage, we employed a fold-wise Min-Max normalization strategy. Scaling parameters were derived exclusively from the training partitions within each cross-validation iteration and subsequently applied to both training and test samples. In detail, we defined $\mathcal{T}_{\text{train}}$ as the training set indices for a given fold. For each feature dimension $j$, we computed the minimum ($\min^{(j)}$) and maximum ($\max^{(j)}$) values solely from $\mathcal{T}_{\text{train}}$. The normalized value $x'_{j}$ for any sample (train or test) was calculated as:
\begin{equation}
x'_{j} = \frac{x_{j} - \min^{(j)}}{\max^{(j)} - \min^{(j)}}.
\end{equation}


\subsection{Deep Learning Architectures }

Using the normalized features, we evaluated three deep learning architectures. The first architecture is the Diff-Logic. The other two architectures are baseline architectures: a multi-layer perceptron (MLP) representing traditional continuous neural networks, and a BNN representing quantized deep learning. The baseline models were used to rigorously evaluate the Diff-Logic.

\subsubsection{Differentiable Logic Gate Network}


\paragraph{ \textbf{Input Quantization: Thermometer Encoding}}
\label{sec:thermo}

Given that the Diff-Logic operates entirely on bitwise logic, the first step was to quantize the normalized features, thus mapping continuous EEG features to binary inputs. To preserve the ordinal relationships of the continuous signal, we employed \textit{Thermometer Encoding} (unary coding). Unlike one-hot encoding, which treats distinct quantization levels as orthogonal, thermometer encoding represents magnitude through cumulative bit activation (e.g., encoding 2 as \texttt{0011} and 3 as \texttt{0111}). This allows the network to exploit the inherent ordering of signal amplitudes, a property shown to help neural networks learn highly non-linear decision boundaries~\cite{buckman2018thermometer}.

The \textit{Thermometer Encoding} discretized the continuous normalized feature $x'_{j}$ into a binary vector of length $T$ using uniformly spaced thresholds. We set $T=15$ because, according to empirical studies~\cite{banner2019post}, this $T$ serves as a practical lower bound for deep learning, preserving task-relevant signals with negligible accuracy loss relative to Float32. For our EEG tasks, this granularity provides necessary signal resolution without excessively inflating the search space. 

\begin{equation}
b_k^{(j)} = \mathbb{I}(x'_{j} \geq \tau_k)
\end{equation}
where $\mathbb{I}(\cdot)$ was the indicator function. For example, if $x'_{j} = 0.6$, the resulting encoding was a sequence of $k$ ones followed by $T-k$ zeros (e.g., $[1, 1, \dots, 1, 0, \dots, 0]$). 

The \textit{Thermometer Encoding} expanded input dimensionality by a factor of $T$: the 19-channel dementia dataset ($95$ features) yielded $1{,}425$ binary inputs, while the 62-channel SEED dataset ($310$ features) yielded $4{,}650$ binary inputs. \\

\paragraph{ \textbf{Differentiable Logic Architecture}}

Unlike standard neural networks that rely on linear transformations followed by non-linear activations, Diff-Logic networks are composed of layers of learnable Boolean logic gates. The core innovation lies in the \textit{soft} logic gate, which allows gradients to flow through discrete Boolean operations during training.

In a Diff-Logic layer, each neuron $n$ connects to two distinct inputs $(x_1, x_2) \in \{0, 1\}^2$. Therefore, there are total of $2^2=4$ combinations (i.e., 00, 01, 10, 11), and as each combination could be mapped to either 0 or 1, there are a total of $2^{2^2} = 16$ possible Boolean functions $f_i: \{0, 1\}^2 \to \{0, 1\}$ (e.g., AND, OR, XOR, NAND). During the training of the Diff-Logic layer, instead of selecting a fixed function initially, the neuron maintains a probability distribution $\boldsymbol{\alpha}_n \in \mathbb{R}^{16}$ over all 16 possible functions (see Table~\ref{app:logic_gates} for relaxations and Figure~\ref{fig:workflow} for illustration). The \textit{soft} output $y_{\text{soft}}$ is the probabilistic sum of all gate outputs:
\begin{equation}
y_{\text{soft}} = \sum_{i=1}^{16} \frac{\exp(\alpha_{n, i})}{\sum_{j} \exp(\alpha_{n,j})} \cdot f_i(x_1, x_2)
\end{equation}
This formulation is fully differentiable with respect to both gate weights $\boldsymbol{\alpha}$ and inputs $\mathbf{x}$, allowing the network to learn through minimizes classification loss via gradient descent algorithm. As training progresses, the distribution $\boldsymbol{\alpha}_n$ sharpens. 

Upon convergence, the network is \textit{hardened} by permanently assigning the single function with the highest probability ($k = \arg\max_i \alpha_{n, i}$) to the neuron. The result is a static, sparse Boolean circuit that executes $y = f_k(x_1, x_2)$ using efficient bitwise operators (AND, OR, NOT, XOR), requiring no floating-point arithmetic.


Our architecture stacked multiple Logic Gate layers (Table ~\ref{tab:architectures}) with random, sparse connectivity. To map the final Boolean outputs to class predictions, we employed a Group Sum operator, where the final layer's bits are partitioned into $C$ groups. The integer logit for class $c$ is the sum of its active bits: $z_c = \sum_{i \in G_c} b_i$. This score was then passed through a temperature-scaled Softmax, defined as $p_c = \exp(z_c / \tau) / \sum_{j=1}^{C} \exp(z_j / \tau)$, to produce class probabilities. Following \cite{petersen2022deep}, we set $\tau=30$ to sufficiently smooth the gradients during training, preventing overconfidence from the discrete integer summations.

\vspace{-0.2cm}

\subsubsection{MLP and BNN models}

The MLP baseline consisted of 2--3 fully connected hidden layers with ReLU activations and dropout ($p=0.3$) after each hidden layer. Crucially, MLP models received the continuous min-max normalized features directly (95 or 310 dimensions), preserving their standard operating conditions.

The BNN baseline followed the architecture of Courbariaux et al.~\cite{courbariaux2016binarized}. Each hidden block consisted of a binarized linear layer, batch normalization, and a sign activation. During the forward pass, weights were binarized to $\{-1, +1\}$ via the sign function, with gradients propagated through a straight-through estimator (STE) that clips the gradient to zero when $|w| > 1$. The latent full-precision weights were retained for gradient updates. Like the MLP, BNN models received continuous normalized inputs.

\section{Experimental Setup}

\subsection{Iso-Parameter Comparison}
\label{sec:iso}

A core tenet of our experimental design was an \textit{iso-parameter comparison}: at each complexity tier, all three architectures were constrained to approximately equal trainable parameter counts, ensuring that performance differences reflect representational efficiency rather than model capacity. MLP and BNN operated on min--max normalized continuous features ($n{=}95$ for dementia, $n{=}310$ for SEED), while Diff-Logic operated on thermometer-encoded binary inputs ($n{=}1{,}425$ and $n{=}4{,}650$, respectively). Hidden layer widths were chosen so that total parameter counts match across architectures at each tier, accounting for BNN's additional BatchNorm parameters per layer. Overall, we evaluated at four tiers: 50k, 100k, 200k, and 500k parameters (Table~\ref{tab:architectures} provides detailed configurations). An exception to the 50k tier was made for the SEED dataset. LogicLayers enforce a structural constraint requiring a minimum of $\lceil n_{\mathrm{in}}/2 \rceil$ outputs per layer. Given the large binary input space for SEED, this constraint establishes a hard lower bound of ${\sim}74$k parameters for the Diff-Logic architecture, dictating the starting tier for this dataset.

\subsection{Training Protocol}
All models were implemented in PyTorch and trained on an Nvidia RTX A6000 using the Adam optimizer for 100 epochs with a batch size of 128. To address class imbalance, we computed inverse-frequency class weights via scikit-learn's \textit{balanced} mode and passed them to the cross-entropy loss function. Furthermore, the Diff-Logic requires a higher learning rate to effectively update the gate probability distributions, so we used $\eta = 0.01$ following Petersen et al.~\cite{petersen2022deep}. MLP and BNN baselines were trained with $\eta = 0.001$. To account for stochasticity in initialization and training dynamics, every experiment was repeated across five independent random seeds. We report results as mean $\pm$ standard deviation. The primary evaluation metric is the \textit{Macro F1-score}, chosen for its robustness to class imbalance compared to standard accuracy.

\subsection{Edge Hardware Benchmarking}
To validate real-time feasibility for portable BCI applications, we benchmarked inference performance on an NVIDIA Jetson Orin Nano (8 GB RAM, ARM Cortex-A78AE CPU, Ampere-architecture GPU, 7W TDP) \cite{NVIDIASuperDevKit2024}. This platform represents the modern embedded systems targeted for wearable neurotechnology \cite{rakhmatulin2023device}. Furthermore, all latency measurements were conducted under single-threaded CPU execution to simulate the worst-case deployment scenario typical of low-power microcontroller environments. Thread counts for PyTorch, OpenMP, and Math Kernel Library (MKL) were explicitly set to one. Each model was evaluated over 10{,}000 timed iterations following a 5{,}000-iteration warm-up phase, using a fixed batch size of 8.

For edge deployment benchmarking, all model architectures were instantiated directly on the Jetson Orin Nano. Diff-Logic models were compiled on-device based on prior work \cite{petersen2022deep}. Because inference latency depends solely on gate count and connectivity, not on which Boolean functions were assigned, untrained instances yield latency measurements identical to those of trained networks, avoiding the need to compile all $5 \times 10 = 50$ trained checkpoints per tier. Boolean input arrays were passed to the compiled shared library via foreign function interface calls, and execution proceeds entirely through ALU bitwise operations without floating-point arithmetic. Finally, MLP and BNN baselines were benchmarked using ONNX Runtime (CPUExecutionProvider, single-threaded) with a batch size of 8 for fair comparison. All measurements use 10,000 timed iterations after a 5,000-iteration warmup. This protocol isolates the raw computational cost of each architecture, independent of parallelism or hardware-specific acceleration.

\begin{table*}[!t]
\centering
\caption{Impact of Model Scale on Classification Performance (Macro F1-Score).}
\label{tab:scale_impact}
\resizebox{\textwidth}{!}{%
\begin{tabular}{@{}ll c | cc | ccc@{}}
\toprule
 & & & \multicolumn{2}{c|}{\textbf{Dementia}} & \multicolumn{3}{c}{\textbf{Emotion (SEED)}} \\ \cmidrule(lr){4-5} \cmidrule(l){6-8} 
\textbf{Model} & \textbf{Type} & \textbf{Params} & \textbf{CN vs AD} & \textbf{CN vs FTD} & \textbf{Chinese} & \textbf{French} & \textbf{German} \\ \midrule

\rowcolor[HTML]{EFEFEF} 
\multicolumn{8}{l}{\textit{Tier 1 (Small Scale)}} \\
$\text{MLP}_{\text{50k}}$ & Float32 & \multirow{3}{*}{\textbf{$\approx$ 50k / 75k}} & 74.8\% $\pm$ 2.6 & 67.4\% $\pm$ 2.8 & \textbf{62.3\% $\pm$ 0.1} & \textbf{60.1\% $\pm$ 1.0} & 54.4\% $\pm$ 1.5 \\
$\text{BNN}_{\text{50k}}$ & Binary & & 55.1\% $\pm$ 7.3 & 59.6\% $\pm$ 12.0 & 49.3\% $\pm$ 2.2 & 55.8\% $\pm$ 2.7 & \textbf{55.4\% $\pm$ 3.5} \\
$\textbf{Diff-Logic}_{\textbf{50k}}$ & \textbf{Logic} & & \textbf{79.5\% $\pm$ 2.1} & \textbf{71.1\% $\pm$ 2.8} & 54.8\% $\pm$ 0.8 & 49.8\% $\pm$ 1.6 & 43.8\% $\pm$ 2.0 \\ \midrule

\rowcolor[HTML]{EFEFEF} 
\multicolumn{8}{l}{\textit{Tier 2 (Medium Scale)}} \\
$\text{MLP}_{\text{100k}}$ & Float32 & \multirow{3}{*}{\textbf{$\approx$ 100k}} & 73.4\% $\pm$ 4.7 & 75.8\% $\pm$ 2.4 & \textbf{62.5\% $\pm$ 0.5} & \textbf{60.2\% $\pm$ 1.6} & 55.2\% $\pm$ 0.6 \\
$\text{BNN}_{\text{100k}}$ & Binary & & 61.2\% $\pm$ 7.3 & 57.7\% $\pm$ 5.8 & 50.1\% $\pm$ 1.1 & 56.3\% $\pm$ 1.3 & \textbf{55.3\% $\pm$ 2.0} \\
$\textbf{Diff-Logic}_{\textbf{100k}}$ & \textbf{Logic} & & \textbf{80.2\% $\pm$ 1.8} & \textbf{79.1\% $\pm$ 2.3} & 55.5\% $\pm$ 0.3 & 50.6\% $\pm$ 1.5 & 45.7\% $\pm$ 2.1 \\ \midrule

\rowcolor[HTML]{EFEFEF} 
\multicolumn{8}{l}{\textit{Tier 3 (Large Scale)}} \\
$\text{MLP}_{\text{200k}}$ & Float32 & \multirow{3}{*}{\textbf{$\approx$ 200k}} & 70.6\% $\pm$ 2.2 & 70.5\% $\pm$ 5.8 & \textbf{61.5\% $\pm$ 0.3} & 56.0\% $\pm$ 3.1 & 50.2\% $\pm$ 2.1 \\
$\text{BNN}_{\text{200k}}$ & Binary & & 58.1\% $\pm$ 7.2 & 58.2\% $\pm$ 3.6 & 48.7\% $\pm$ 3.2 & \textbf{57.9\% $\pm$ 1.9} & \textbf{56.2\% $\pm$ 2.9} \\
$\textbf{Diff-Logic}_{\textbf{200k}}$ & \textbf{Logic} & & \textbf{80.0\% $\pm$ 3.7} & \textbf{76.4\% $\pm$ 2.8} & 56.4\% $\pm$ 0.7 & 52.9\% $\pm$ 0.8 & 50.2\% $\pm$ 1.3 \\ \midrule

\rowcolor[HTML]{EFEFEF} 
\multicolumn{8}{l}{\textit{Tier 4 (X-Large Scale)}} \\
$\text{MLP}_{\text{500k}}$ & Float32 & \multirow{3}{*}{\textbf{$\approx$ 500k}} & 72.0\% $\pm$ 5.1 & 69.8\% $\pm$ 5.7 & \textbf{61.3\% $\pm$ 0.4} & 56.6\% $\pm$ 2.3 & 46.6\% $\pm$ 2.8 \\
$\text{BNN}_{\text{500k}}$ & Binary & & 59.3\% $\pm$ 11.0 & 56.0\% $\pm$ 3.3 & 52.3\% $\pm$ 0.6 & \textbf{57.9\% $\pm$ 0.7} & \textbf{54.6\% $\pm$ 2.2} \\ 
$\textbf{Diff-Logic}_{\textbf{500k}}$ & \textbf{Logic} & & \textbf{78.9\% $\pm$ 0.7} & \textbf{74.1\% $\pm$ 2.2} & 61.0\% $\pm$ 0.5 & 54.7\% $\pm$ 1.3 & 50.7\% $\pm$ 2.0 \\ \bottomrule
\end{tabular}%
}
\footnotesize{\textit{Note:} Chance-level performance for the binary dementia dataset is 50\%, while the 3-class SEED datasets have a chance level of 33.3\%. Parameter count 50k for dementia and 74k for SEED (Diff-Logic minimum)}
\end{table*}


\begin{table*}
\centering
\caption{Inference Speed on Nvidia Jetson Orin Nano with Single-core CPU at 960 MHz.}
\label{tab:inference_efficiency_7w_combined}
\resizebox{\textwidth}{!}{%
\begin{tabular}{@{}ll c | ccc | ccc@{}}
\toprule
 & & & \multicolumn{3}{c|}{\textbf{Dementia}} & \multicolumn{3}{c}{\textbf{Emotion (SEED)}} \\ \cmidrule(lr){4-6} \cmidrule(l){7-9} 
\textbf{Model} & \textbf{Runtime} & \textbf{Size (KB)} & \textbf{F1-Score} & \textbf{Latency} & \textbf{Speedup} & \textbf{F1-Score} & \textbf{Latency} & \textbf{Speedup} \\ \midrule

\rowcolor[HTML]{EFEFEF} 
\multicolumn{9}{l}{\textit{Tier 1 (Small Scale)}} \\
MLP & ONNX (Float32) & $\approx$195 & 74.8\% & \textbf{0.107 ms} & 1.00$\times$ & \textbf{62.3\%} & \textbf{0.136 ms} & 1.00$\times$ \\
BNN & ONNX (Binary) & $\approx$198 & 55.1\% & 0.178 ms & 0.61$\times$ & 49.3\% & 0.192 ms & 0.70$\times$ \\ 
\textbf{Diff-Logic} & \textbf{C-Compiled} & \textbf{$\approx$52} & \textbf{79.5\%} & 0.193 ms & 0.56$\times$ & 54.8\% & 0.223 ms & 0.61$\times$ \\ \midrule

\rowcolor[HTML]{EFEFEF} 
\multicolumn{9}{l}{\textit{Tier 2 (Medium Scale)}} \\
MLP & ONNX (Float32) & $\approx$389 & 73.4\% & \textbf{0.167 ms} & 1.00$\times$ & \textbf{62.5\%} & \textbf{0.165 ms} & 1.00$\times$ \\
BNN & ONNX (Binary) & $\approx$394 & 61.2\% & 0.268 ms & 0.62$\times$ & 50.1\% & 0.231 ms & 0.71$\times$ \\ 
\textbf{Diff-Logic} & \textbf{C-Compiled} & \textbf{$\approx$44} & \textbf{80.2\%} & 0.191 ms & 0.87$\times$ & 55.5\% & 0.224 ms & 0.74$\times$ \\ \midrule

\rowcolor[HTML]{EFEFEF} 
\multicolumn{9}{l}{\textit{Tier 3 (Large Scale) -- The Efficiency Crossover}} \\
MLP & ONNX (Float32) & $\approx$779 & 70.6\% & 0.277 ms & 1.00$\times$ & \textbf{61.5\%} & 0.279 ms & 1.00$\times$ \\
BNN & ONNX (Binary) & $\approx$781 & 58.1\% & 0.426 ms & 0.65$\times$ & 48.7\% & 0.382 ms & 0.73$\times$ \\ 
\textbf{Diff-Logic} & \textbf{C-Compiled} & \textbf{$\approx$64} & \textbf{80.0\%} & \textbf{0.191 ms} & \textbf{1.45$\times$} & 56.4\% & \textbf{0.265 ms} & \textbf{1.05$\times$} \\ \midrule

\rowcolor[HTML]{EFEFEF} 
\multicolumn{9}{l}{\textit{Tier 4 (X-Large Scale)}} \\
MLP & ONNX (Float32) & $\approx$1959 & 72.0\% & 0.631 ms & 1.00$\times$ & \textbf{61.3\%} & 0.625 ms & 1.00$\times$ \\
BNN & ONNX (Binary) & $\approx$1963 & 59.3\% & 0.939 ms & 0.67$\times$ & 52.3\% & 0.819 ms & 0.76$\times$ \\ 
\textbf{Diff-Logic} & \textbf{C-Compiled} & \textbf{$\approx$140} & \textbf{78.9\%} & \textbf{0.216 ms} & \textbf{2.91$\times$} & 61.0\% & \textbf{0.271 ms} & \textbf{2.30$\times$} \\ \bottomrule
\end{tabular}%
}
\end{table*}

\section{Experimental Results}

\subsection{Classification Performance and Robustness}

As shown in Table~\ref{tab:scale_impact}, Diff-Logic demonstrated superior performance on the dementia classification tasks, consistently outperforming both the MLP (Float32) and BNN (Binary) baselines. Specifically, on the CN vs.\ AD task, Diff-Logic achieved an F1-score of 78.9--80.2\% across all model scales, representing a significant improvement of nearly 10 percentage points over the MLP baseline at the 200k tier. In contrast, the BNN baseline exhibited high instability, particularly on the dementia tasks, where standard deviations reached $\pm$12. This suggests that the gradient approximation methods required for training standard binary neural networks struggled to converge on these high-dimensional EEG data. Diff-Logic, by learning connectivity and logic directly, appeared to avoid these optimization pitfalls, offering a more robust alternative for quantized inference.

On the SEED emotion recognition benchmarks, the Float32 MLP maintained a moderate performance lead ($\approx$62 vs 55--61). However, it is crucial to note that this accuracy gap \textbf{vanished} at the 500k parameter tier (Diff-Logic 61.0 vs. MLP 61.3). This indicates that while Diff-Logic requires slightly higher capacity to model the complex emotional features in SEED, it can match full-precision performance when scaled, without the associated latency penalties discussed below.

\subsection{Inference Efficiency and Scaling Laws}

\begin{figure*}[h]
    \centering
    \noindent 
    \begin{subfigure}[b]{0.45\textwidth}
        \centering
        \includegraphics[width=\linewidth]{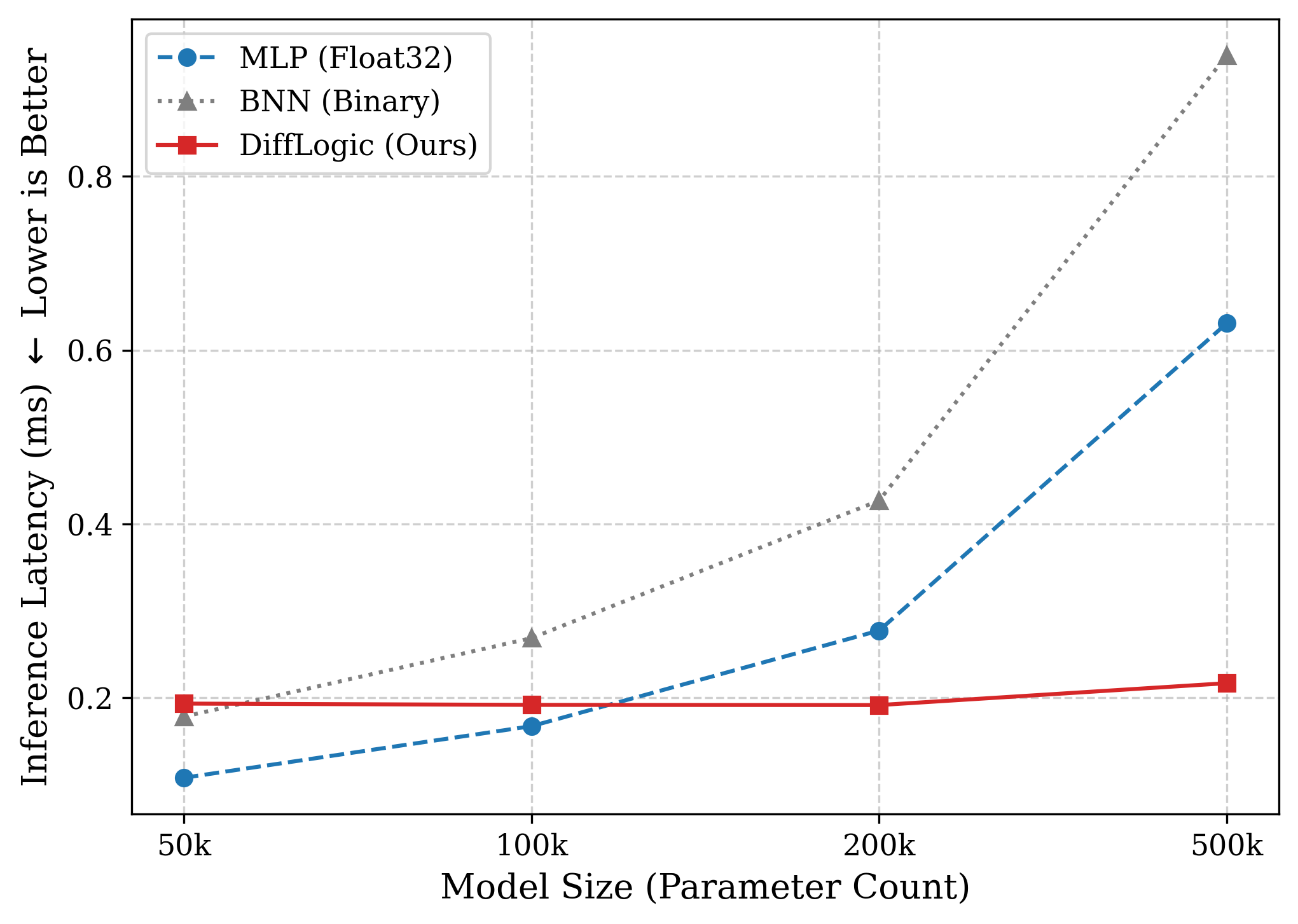}
        \caption{ALZ: Inference Latency vs. Model Scale}
        \label{fig:alz_latency}
    \end{subfigure}
    \begin{subfigure}[b]{0.45\textwidth}
        \centering
        \includegraphics[width=\linewidth]{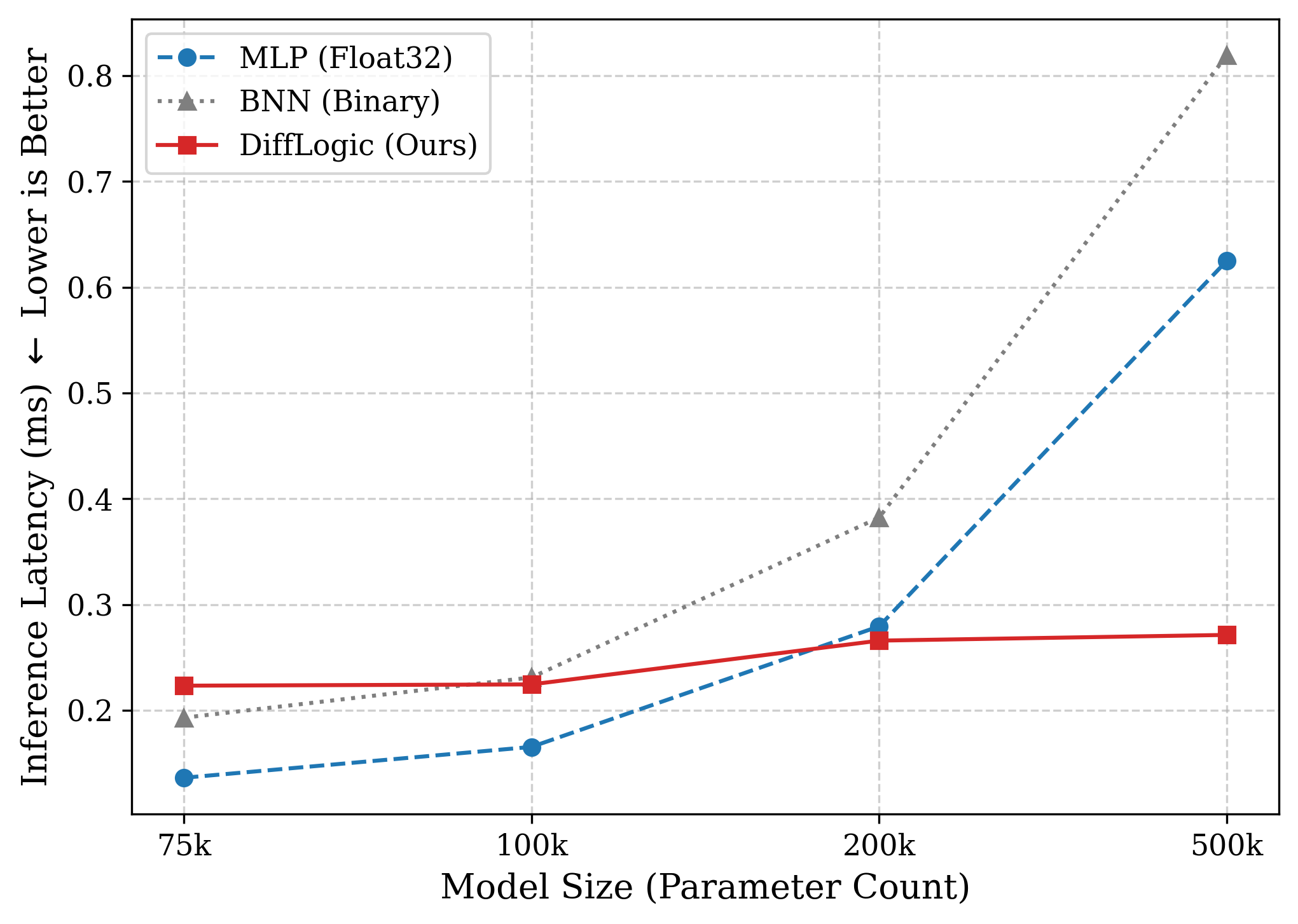}
        \caption{SEED: Inference Latency vs. Model Scale}
        \label{fig:seed_latency}
    \end{subfigure}
    
    \vspace{0.3cm}
    
    \noindent 
    \begin{subfigure}[b]{0.45\textwidth}
        \centering
        \includegraphics[width=\linewidth]{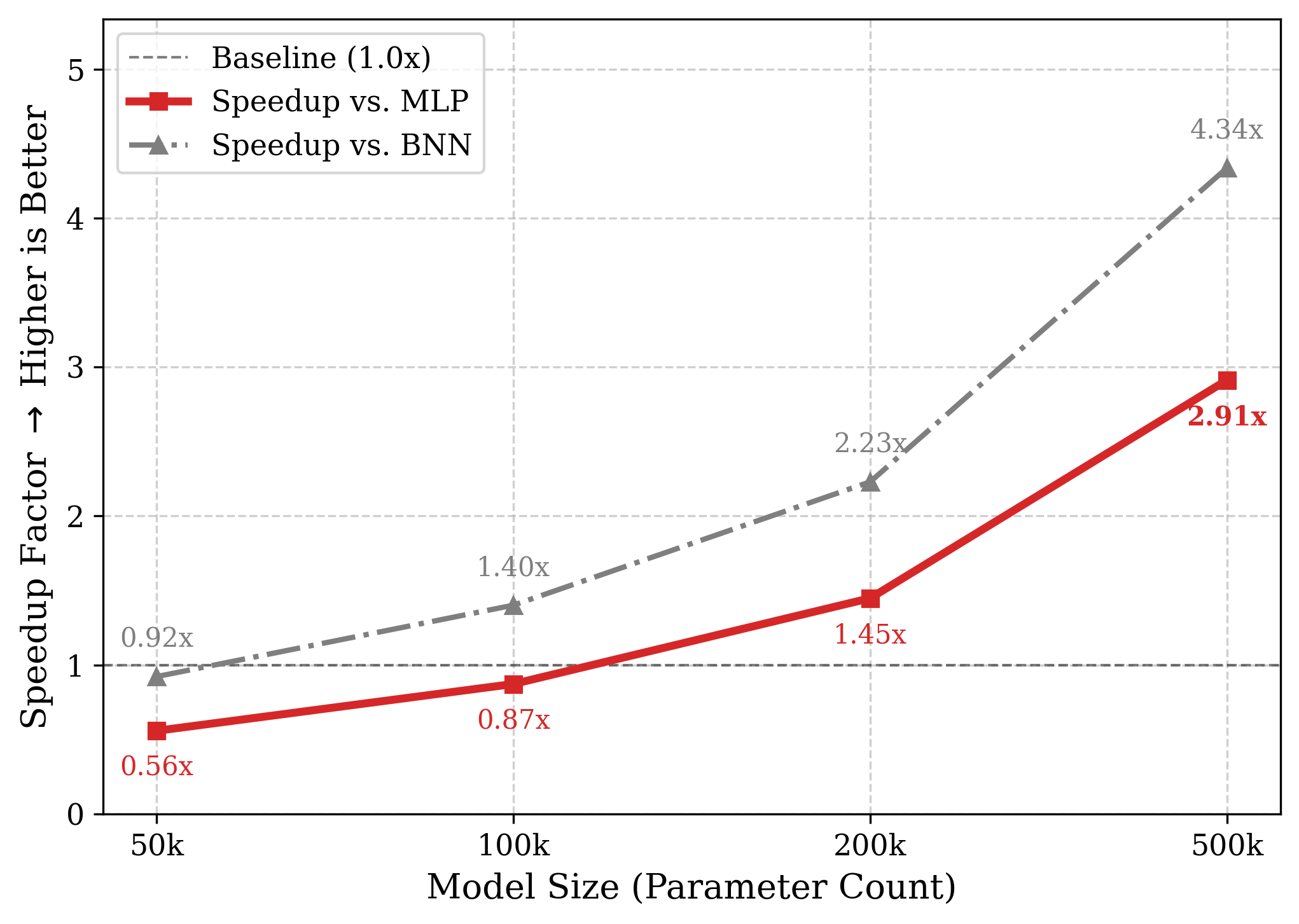}
        \caption{ALZ: Speedup (Diff-Logic vs. Baselines)}
        \label{fig:alz_speedup}
    \end{subfigure}
    \begin{subfigure}[b]{0.45\textwidth}
        \centering
        \includegraphics[width=\linewidth]{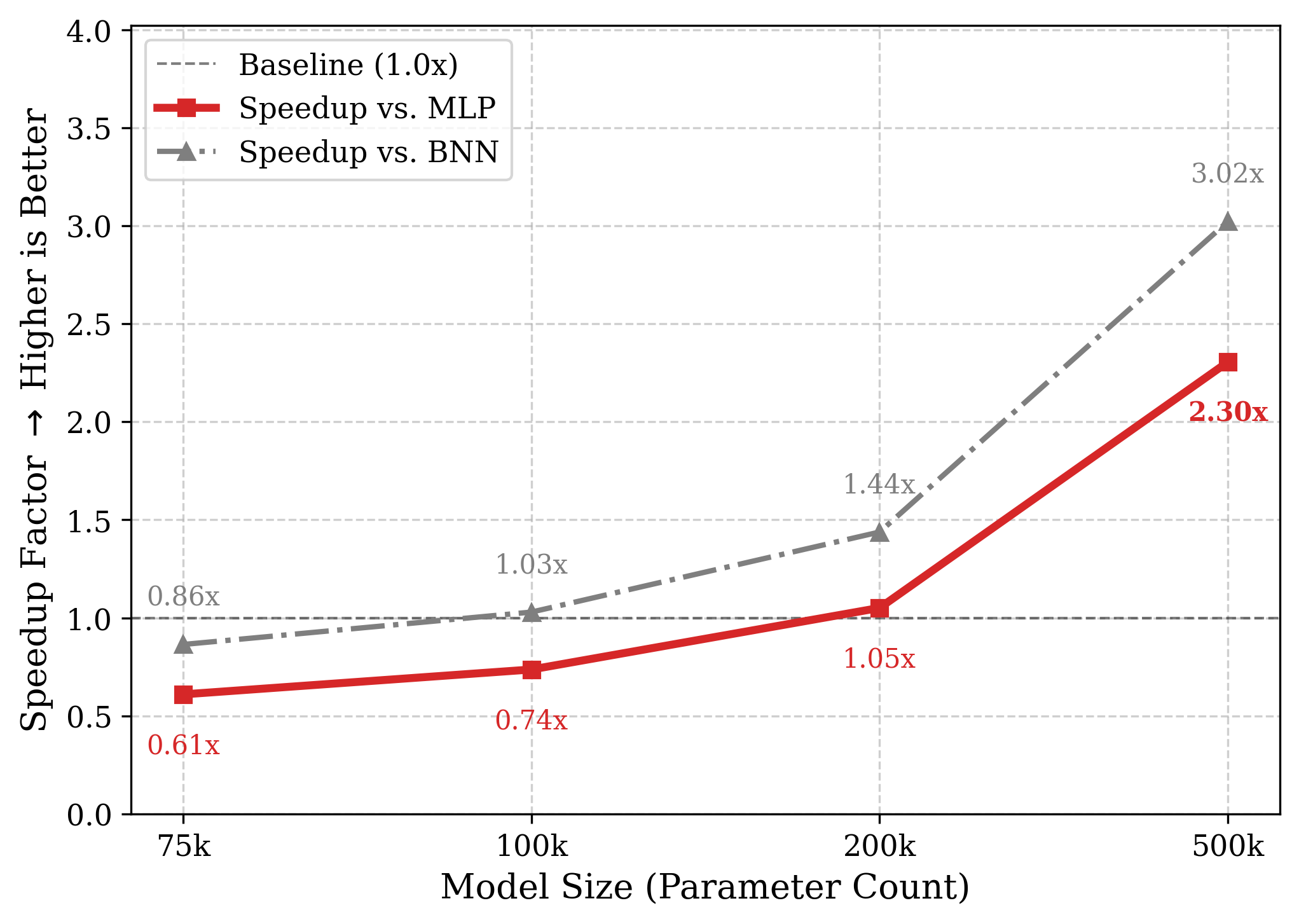}
        \caption{SEED: Speedup (Diff-Logic vs. Baselines)}
        \label{fig:seed_speedup}
    \end{subfigure}
    
    \caption{Inference scaling on the Jetson Orin Nano embedded CPU (single-threaded).}
    \label{fig:scaling_analysis}
\end{figure*}

The most critical advantage of the proposed method was revealed in the scaling behaviour (Figure~\ref{fig:scaling_analysis}). While the inference latency of the MLP and BNN baselines scaled linearly with parameter count, Diff-Logic latency remained effectively constant (${\sim}0.19$--$0.22$ms) even as the training model size grew by an order of magnitude. This efficiency, and the resulting \textit{flat} scaling curve, stems from two key properties of the Diff-Logic architecture. First, through \textit{structural condensation}, the training formulation explored a dense search space (e.g., 16 operators per neuron), but the inference model collapsed to a single hard logic gate per node. This effectively reduced the active parameter count by a factor of 16$\times$ relative to the training capacity. Second, \textit{bitwise compilation} translated these sparse logic circuits into flat bitwise operations. This allowed the model to leverage the full width of CPU registers (SWAR) and Single Instruction, Multiple Data (SIMD) instructions, minimizing memory access bottlenecks. Consequently, at the 500k tier, Diff-Logic achieved a 2.91$\times$ speedup over the MLP and a 4.33$\times$ speedup over the BNN. Notably, Diff-Logic outperformed the BNN despite both using binary operations, highlighting that logic gate sparsity is fundamentally more hardware-efficient than simple weight quantization. 

\begin{figure*}[t]
    \centering
    \begin{subfigure}[b]{0.45\textwidth}
        \centering
        \includegraphics[width=\textwidth]{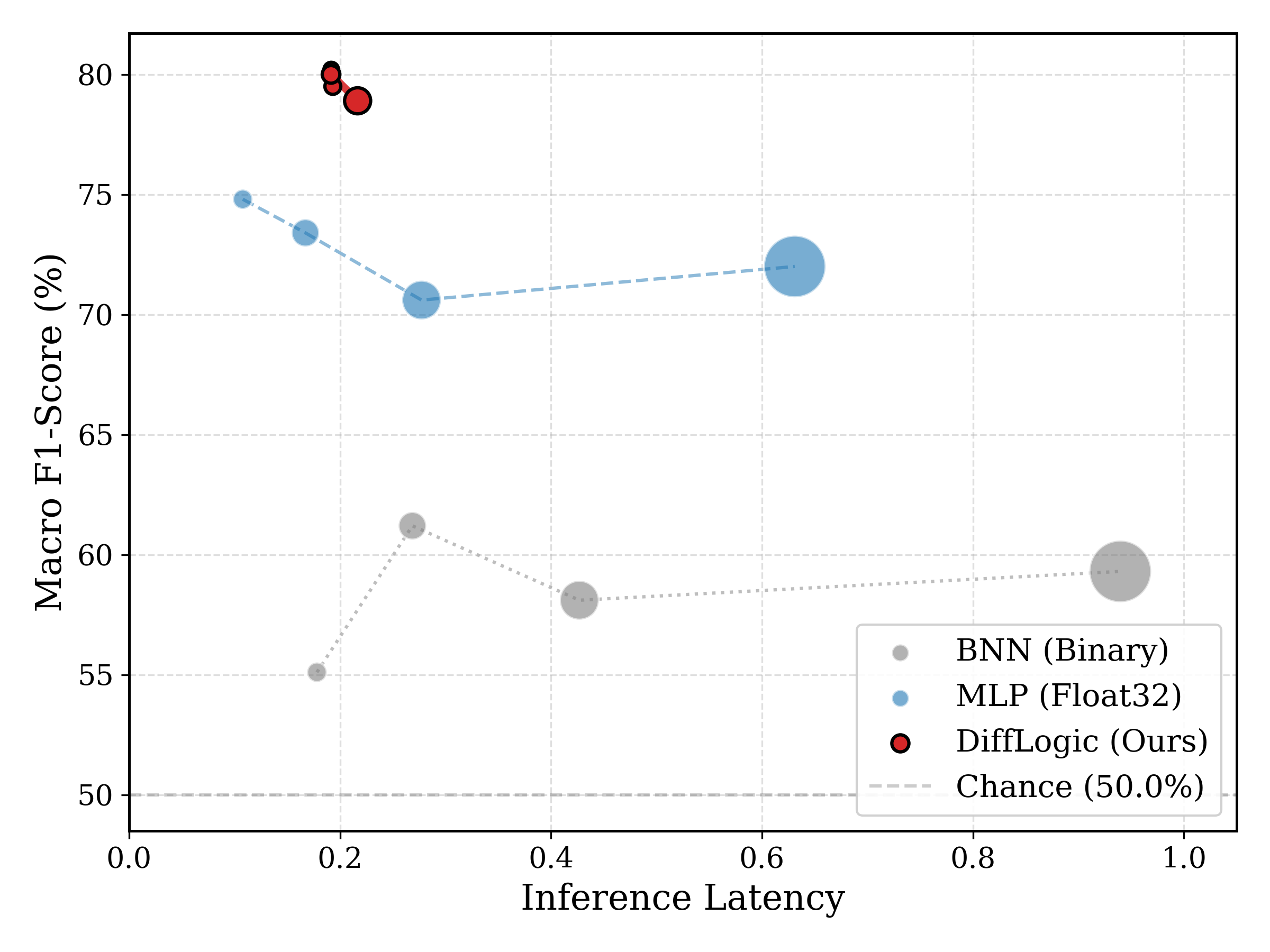}
        \caption{Dementia (N vs.\ AD)}
        \label{fig:f1_latency_alz}
    \end{subfigure}
    \begin{subfigure}[b]{0.45\textwidth}
        \centering
        \includegraphics[width=\textwidth]{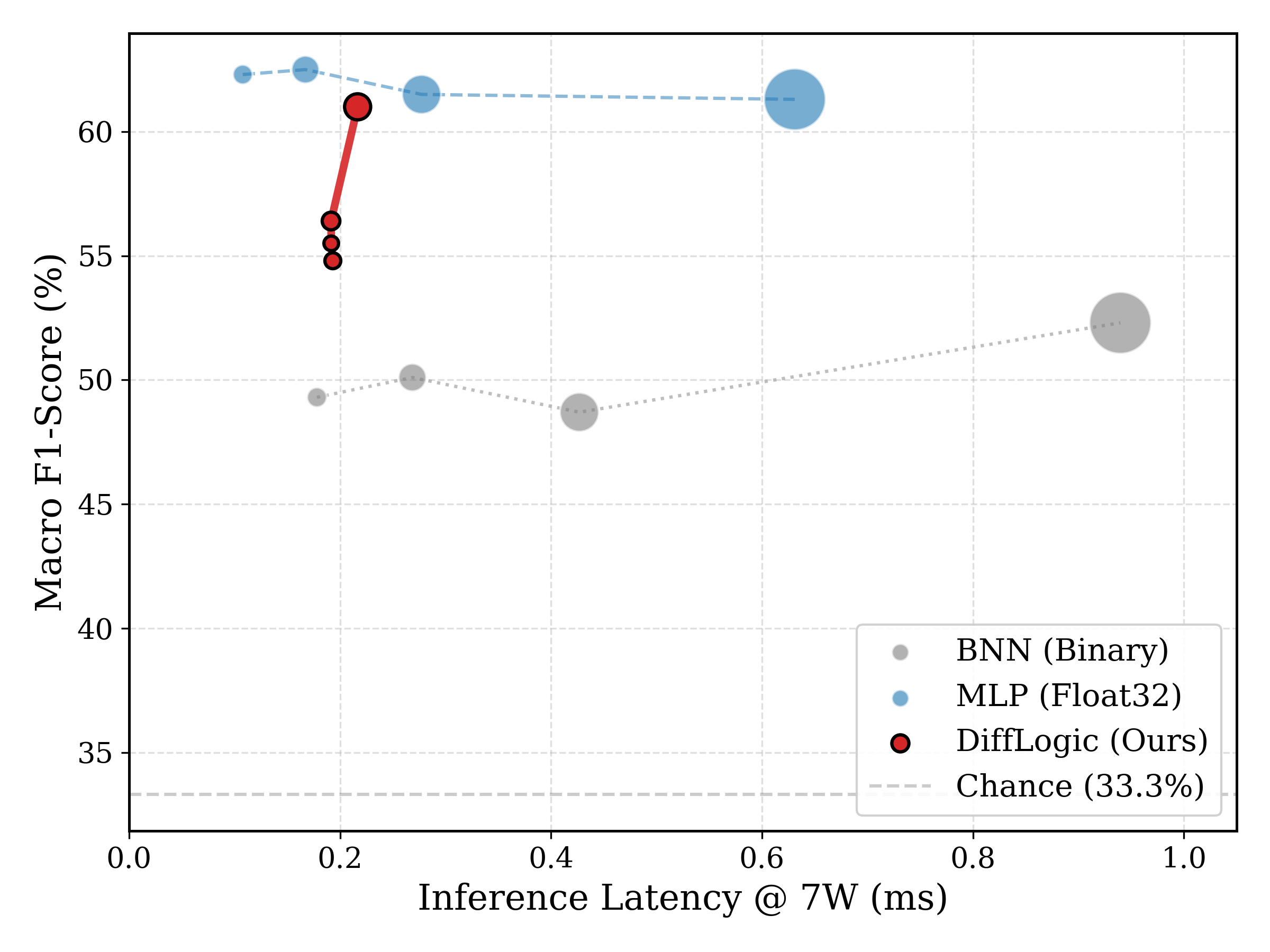}
        \caption{SEED (3-class Emotion)}
        \label{fig:f1_latency_seed}
    \end{subfigure}
    \caption{Macro F1-Score vs.\ Inference Latency on the Jetson Orin Nano embedded CPU (single-threaded). Each point represents one model tier (50k--500k parameters) and bubble area is proportional to compiled model storage size in KB.}
    \vspace{-0.2cm}
    \label{fig:f1_vs_latency}
\end{figure*}

Finally, Figure~\ref{fig:f1_vs_latency} summarizes the performance--efficiency trade-off. On the dementia task (Figure~\ref{fig:f1_latency_alz}), Diff-Logic dominated the pareto frontier, simultaneously achieving the highest F1-score and the lowest latency at scale, a combination unmatched by either baseline. On the SEED task (Figure~\ref{fig:f1_latency_seed}), while the Float32 MLP retained a slight peak performance advantage, it did so at a disproportionate computational and spatial cost. Specifically, at the 500k parameter tier, the MLP required nearly 2 MB of storage and 0.63 ms per inference. In stark contrast, the equivalent Diff-Logic model completed the forward pass in just 0.22 ms and consumed only 140 KB of memory, representing a substantial 14$\times$ reduction in storage footprint. For constrained embedded environments like the Jetson Orin Nano, where L1/L2 cache sizes are critical bottlenecks, this dramatic reduction in footprint and latency suggests that Diff-Logic is a highly viable candidate for always-on deployment, easily justifying the minor trade-off in peak performance observed in more complex datasets (SEEDs).

\section{Discussion} 
\label{sec:discussion}

\subsection{Discrete Logic as an Effective Inductive Bias for EEG}
Our results indicate that Diff-logic matches or exceeds deep learning performance on EEG tasks while using significantly fewer resources. Notably, Diff-Logic outperformed the MLP on dementia tasks by 9.4 percentage points in F1-score with identical parameter budgets (Tier 3), suggesting a superior inductive bias for clinical signals. While MLPs must approximate sharp decision boundaries with continuous activations, Diff-Logic naturally aligns with the threshold-based rules common in clinical EEG biomarkers (e.g., specific band-power cutoffs). 

Scaling analysis revealed no benefit from increasing parameters beyond 100k (Table~\ref{tab:scale_impact}). In fact, MLP accuracy on the CN vs.\ AD task dropped from 74.8\% to 72.0\% when scaled from 50k to 500k, indicating overfitting. This highlights again the practical value of logic-based models in clinical settings with smaller cohorts, where they offer superior generalization alongside deployment efficiency.

The BNN baseline consistently lagged on dementia tasks, likely due to the restricted capacity caused by binarizing both weights and activations. However, this extreme quantization acted as an implicit regularizer at higher parameter counts.


While the Float32 MLP degraded by nearly 8 percentage points on SEED German tasks when scaled to 500k, the BNN maintained stable performance across all scales (Table~\ref{tab:scale_impact}, SEED German) and even overtook the MLP on larger-scale SEED French tasks, effectively avoiding catastrophic overfitting. Finally, regarding inference efficiency, Diff-Logic retains a fundamental architectural advantage over BNNs even when discounting implementation overheads. While BNNs require an accumulation step (\textit{popcount}) after the $XNOR$ operation to sum bitwise products, compiled Diff-Logic models are accumulation-free, operating as discrete combinational circuits.

\subsection{The scaling advantage of Boolean circuits}
The most striking efficiency result was not Diff-Logic's absolute speed, but its \textit{scaling behaviour}. The inference latency of the MLP and BNN baselines exhibits an exponential behaviour as parameter count increases. In contrast, compiled Diff-Logic latency remained nearly constant across a 10$\times$ increase in model size (Figure~\ref{fig:scaling_analysis}). This occurred because the compiled Boolean circuit is evaluated via bitwise operations over packed registers. Consequently, the computational cost was bounded by the circuit's depth (the number of sequential layers) rather than its width (the number of parallel gates). At the 500k parameter tier, this architectural advantage yielded a 2.91$\times$ speedup over the MLP on a single ARM core.

\subsection{Implications for edge deployment}
Real-time clinical tasks like seizure detection require ultra-low latency to track neural dynamics on portable, battery-powered devices. At the 500k tier, Diff-Logic achieved a latency of 0.22\, ms with a total storage footprint of just 140 KB. This was nearly two orders of magnitude below the real-time threshold and compact enough to fit entirely within the L1/L2 cache hierarchy of most modern microcontrollers. This effectively eliminated costly off-chip memory accesses, positioning logic-based networks as a highly viable architecture class for always-on neural monitoring, where the dominant engineering constraints are power and memory rather than raw continuous accuracy.

\subsection{Limitations and Future Work} 
A primary limitation of our current evaluation is its reliance on pre-extracted tabular EEG features. The standard Diff-Logic architecture does not inherently support temporal or spatial convolutions, restricting its immediate applicability to raw time-series or image-based modalities. Furthermore, the thermometer encoding expanded the input dimensionality by 15$\times$, which may become computationally expensive for very high-dimensional raw signals. Nevertheless, for real-time applications, our single-core results remain highly representative of true edge deployment constraints.

Several directions merit further investigation to bridge these gaps. First, recent extensions of Diff-Logic to convolutional~\cite{petersen2024convolutional} and recurrent~\cite{buhrer2025recurrent} gate architectures offer a path toward temporal and spatial EEG feature extraction using logic primitives. Integrating these variants could enable end-to-end learning directly from raw EEG while preserving the accumulation-free inference efficiency demonstrated in this work. Second, for logic-based networks to reach true clinical viability, robust interpretability methods must be developed. Although we can currently observe gate-type distributions and node-selection frequencies per layer, designing systematic approaches to extract and validate clinically meaningful Boolean decision rules from these compiled circuits remains an open challenge. Finally, true Field-Programmable Gate Array (FPGA) or Application-Specific Integrated Circuit (ASIC) implementations of the compiled Boolean circuits, where logic gates map directly to silicon primitives without software emulation, could bring Diff-Logic substantially closer to ultra-low-power, real-time brain--computer interface applications.

\section{Conclusion}
We presented an iso-parameter comparison of Diff-Logic against standard continuous (MLP) and binary (BNN) baselines for EEG classification. Our experiments show that Diff-Logic achieved superior performance on clinical dementia detection (up to 80.2\% Macro F1, outperforming the MLP by 9.4 percentage points) and comparable accuracy on the SEED emotion recognition task at scale. Crucially, these logic networks compiled to accumulation-free Boolean circuits that were 2.91$\times$ faster and 14$\times$ smaller than equivalent floating-point models at inference time. Furthermore, we showed that Diff-Logic's latency remained effectively constant as model complexity grew, a hardware scaling property unmatched by conventional multiply-accumulate architectures. Ultimately, these results demonstrate that logic-based neural networks offer a compelling performance--efficiency trade-off for resource-constrained biomedical applications. Our findings emphasize that architectural inductive bias is a decisive factor for embedded neural inference, positioning Diff-logic as a deployment-ready alternative to standard deep learning for structured biomedical signals.

\section*{Acknowledgements}
 This work was supported by NSERC Discovery Grants (RGPIN-2023-03443 and RGPIN-2024-05575) and the Manitoba Medical Service Foundation (grant number 2026-04).

\printbibliography[heading=subbibintoc]

\newpage


\newpage
\appendix
\renewcommand{\thetable}{A\arabic{table}}
\setcounter{table}{0}
\renewcommand{\thefigure}{A\arabic{figure}}
\setcounter{figure}{0}

\section{Dataset Summary}
\label{app:datasets}

\begin{table}[h]
\centering
\caption{Summary of EEG datasets. Features denote Power Spectral Density (PSD) or Differential Entropy (DE). Windows denotes total non-overlapping segments across all subjects.}
\label{tab:datasets}
\resizebox{\columnwidth}{!}{%
\begin{tabular}{@{}lccccccc@{}}
\toprule
\textbf{Dataset} & \textbf{Type} & \textbf{Subj.} & \textbf{Chan.} & \textbf{$f_s$ (Hz)} & \textbf{Feat.\ Dim} & \textbf{Windows} & \textbf{CV Protocol} \\ \midrule
Dementia (N vs AD) & Clinical & 65 & 19 & 500 & 95 (PSD) & 5{,}292 & Subject-level \\
Dementia (N vs FTD) & Clinical & 52 & 19 & 500 & 95 (PSD) & 4{,}043 & Subject-level \\ \midrule
SEED-China & Emotion & 15 & 62 & 200 & 310 (DE) & 37{,}890 & Trial-level \\
SEED-France & Emotion & 8 & 62 & 200 & 310 (DE) & 20{,}016 & Trial-level \\
SEED-Germany & Emotion & 8 & 62 & 200 & 310 (DE) & 17{,}420 & Trial-level \\ \bottomrule
\end{tabular}%
}
\end{table}


\section{Thermometer Encoding}
\label{app:thermo}

\begin{table}[h]
\centering
\caption{Mapping between continuous input ranges, Thermometer Encoding ($T=15$), and equivalent 4-bit unsigned integer precision.}
\label{tab:quantization_map}
\resizebox{0.9\columnwidth}{!}{%
\begin{tabular}{@{}cccc@{}}
\toprule
\textbf{Normalized Input} $x'$ & \textbf{Thermometer Code} (15-bit) & \textbf{Sum of Bits} & \textbf{UINT4 Equiv.} \\ \midrule
$[0, \frac{1}{16})$            & $[0, 0, 0, \dots, 0]$ & 0  & 0000 (0) \\
$[\frac{1}{16}, \frac{2}{16})$ & $[1, 0, 0, \dots, 0]$ & 1  & 0001 (1) \\
$[\frac{2}{16}, \frac{3}{16})$ & $[1, 1, 0, \dots, 0]$ & 2  & 0010 (2) \\
\vdots                         & \vdots                 & \vdots & \vdots \\
$[\frac{15}{16}, 1]$           & $[1, 1, 1, \dots, 1]$ & 15 & 1111 (15) \\ \bottomrule
\end{tabular}%
}
\end{table}

\newpage
\section{Boolean operators for Differential Logic Gate Network}
\label{app:datasets}

\begin{table}[h]
    \centering
    \caption{The 16 Boolean logic gates and their differentiable real-valued relaxations used in Diff-Logic. The inputs $a, b \in [0, 1]$ represent the probabilities of the input bits being true.}
    \label{app:logic_gates}
    \resizebox{0.95\linewidth}{!}{ 
    \begin{tabular}{clll}
        \toprule
        \textbf{ID} & \textbf{Boolean Operator} & \textbf{Description} & \textbf{Real-Valued Relaxation} $f(a, b)$ \\
        \midrule
        0 & $\textbf{False}$ & Constant False & $0$ \\
        1 & $a \land b$ & AND & $a \cdot b$ \\
        2 & $a \land \neg b$ & A and not B & $a - a \cdot b$ \\
        3 & $a$ & Identity A & $a$ \\
        4 & $\neg a \land b$ & Not A and B & $b - a \cdot b$ \\
        5 & $b$ & Identity B & $b$ \\
        6 & $a \oplus b$ & XOR & $a + b - 2(a \cdot b)$ \\
        7 & $a \lor b$ & OR & $a + b - (a \cdot b)$ \\
        8 & $\neg(a \lor b)$ & NOR & $1 - (a + b - a \cdot b)$ \\
        9 & $\neg(a \oplus b)$ & XNOR & $1 - (a + b - 2a \cdot b)$ \\
        10 & $\neg b$ & NOT B & $1 - b$ \\
        11 & $a \lor \neg b$ & Implication (B $\to$ A) & $1 - b + a \cdot b$ \\
        12 & $\neg a$ & NOT A & $1 - a$ \\
        13 & $\neg a \lor b$ & Implication (A $\to$ B) & $1 - a + a \cdot b$ \\
        14 & $\neg(a \land b)$ & NAND & $1 - a \cdot b$ \\
        15 & $\textbf{True}$ & Constant True & $1$ \\
        \bottomrule
    \end{tabular}
    }
\end{table}

\section{Model Architectures Structure for Comparison}
\label{app:thermo}

\begin{table}[h]
\centering
\caption{Model architectures at each iso-parameter tier. Diff-Logic uses $H$ gates/layer; MLP and BNN use $H$ neurons/layer. Diff-Logic receives thermometer-encoded binary inputs; MLP and BNN receive continuous inputs.}
\label{tab:architectures}
\resizebox{\columnwidth}{!}{%
\begin{tabular}{@{}llll@{}}
\toprule
\textbf{Tier} & \textbf{Diff-Logic (Binary In)} & \textbf{MLP (Continuous In)} & \textbf{BNN (Continuous In)} \\ \midrule
\multicolumn{4}{@{}l}{\textit{Dementia (input: 1{,}425 binary / 95 continuous, 2 classes)}} \\
50k  & 2L: [1532, 1532]               & 2L: [310, 64]            & 2L: [306, 64]   \\
100k & 4L: [1940, 1940, 970, 970]     & 2L: [620, 64]            & 2L: [616, 64]   \\
200k & 6L: [2988$\times$2, 1992$\times$2, 996$\times$2]  & 3L: [870, 128, 32]  & 3L: [860, 128, 32] \\
500k & 6L: [7782$\times$2, 5188$\times$2, 2594$\times$2] & 3L: [2200, 128, 64] & 3L: [2180, 128, 64] \\ \midrule
\multicolumn{4}{@{}l}{\textit{SEED (input: 4{,}650 binary / 310 continuous, 3 classes)}} \\
74k  & 2L: [2325, 2325]               & 2L: [216, 32]            & 2L: [214, 32]   \\
100k & 2L: [3219, 3219]               & 2L: [292, 32]            & 2L: [290, 32]   \\
200k & 2L: [6156, 6156]               & 3L: [530, 64, 32]       & 3L: [520, 64, 32] \\
500k & 4L: [10314$\times$2, 5157$\times$2] & 3L: [1120, 128, 64] & 3L: [1120, 128, 64] \\ \bottomrule
\end{tabular}%
}
\end{table}

\section{Exact Model Configurations}
\label{app:configs}

\begin{table}[h]
\centering
\caption{Exact parameter and gate counts. For MLP and BNN, inference cost scales with the full parameter count. For Diff-Logic, inference cost scales only with the compiled gate count.}
\label{tab:exact_params}
\begin{tabular}{@{}l rr rr@{}}
\toprule
& \multicolumn{2}{c}{\textbf{Dementia}} & \multicolumn{2}{c}{\textbf{SEED}} \\
\cmidrule(lr){2-3} \cmidrule(lr){4-5}
\textbf{Model} & \textbf{Params} & \textbf{Gates} & \textbf{Params} & \textbf{Gates} \\ \midrule
MLP$_{\text{50k}}$       & 49{,}794  & ---    & 74{,}219  & ---    \\
BNN$_{\text{50k}}$       & 49{,}894  & ---    & 74{,}025  & ---    \\
Diff-Logic$_{\text{50k}}$ & 49{,}024  & 3{,}064  & 74{,}400  & 4{,}650  \\ \midrule
MLP$_{\text{100k}}$       & 99{,}394  & ---    & 100{,}287 & ---    \\
BNN$_{\text{100k}}$       & 100{,}114 & ---    & 100{,}245 & ---    \\
Diff-Logic$_{\text{100k}}$ & 93{,}120  & 5{,}820  & 103{,}008 & 6{,}438  \\ \midrule
MLP$_{\text{200k}}$       & 199{,}202 & ---    & 200{,}993 & ---    \\
BNN$_{\text{200k}}$       & 199{,}002 & ---    & 198{,}475 & ---    \\
Diff-Logic$_{\text{200k}}$ & 191{,}232 & 11{,}952 & 196{,}992 & 12{,}312 \\ \midrule
MLP$_{\text{500k}}$       & 501{,}314 & ---    & 500{,}259 & ---    \\
BNN$_{\text{500k}}$       & 501{,}578 & ---    & 502{,}883 & ---    \\
Diff-Logic$_{\text{500k}}$ & 498{,}048 & 31{,}128 & 495{,}072 & 30{,}942 \\ \bottomrule
\end{tabular}%

\end{table}

\end{document}